\documentclass[runningheads]{llncs}
\usepackage{graphicx}
\usepackage{comment}
\usepackage{amsmath,amssymb} 
\usepackage{color}

\usepackage[width=122mm,left=12mm,paperwidth=146mm,height=193mm,top=12mm,paperheight=217mm]{geometry}

\usepackage[utf8]{inputenc}
\usepackage{bm}
\usepackage{mathrsfs}
\usepackage{enumitem}
\usepackage{multirow}
\usepackage{caption2}
\usepackage{booktabs}
\usepackage{amsfonts}
\usepackage{float}
\usepackage{lipsum}
\usepackage{array}
\usepackage{arydshln}
\usepackage{subfigure}

\begin{document}
\pagestyle{headings}
\mainmatter
\def\ECCVSubNumber{1881}  

\title{Super Resolution Using Segmentation-Prior Self-Attention Generative Adversarial Network} 


\titlerunning{Super Resolution Using SPSAGAN}
%
\author{Yuxin Zhang\inst{1} \and
Zuquan Zheng\inst{2} \and
Roland Hu\thanks{Contact author}\inst{1}}
\authorrunning{Zhang et al.}
%
\institute{
Zhejiang University, China \\
\email{\{yuxinzhang, haoji\_hu\}@zju.edu.cn} \and
Central Media Technology Institute, Huawei Technologies Co., Ltd, China \\
\email{zhengzuquan@huawei.com}}
\maketitle

\begin{abstract}
Convolutional Neural Network~(CNN) is intensively implemented to solve super resolution~(SR) tasks because of its superior performance. However, the problem of super resolution is still challenging due to the lack of prior knowledge and small receptive field of CNN. We propose the Segmentation-Piror Self-Attention Generative Adversarial Network~(SPSAGAN) to combine segmentation-priors and feature attentions into a unified framework. This combination is led by a carefully designed weighted addition to balance the influence of feature and segmentation attentions, so that the network can emphasize textures in the same segmentation category and meanwhile focus on the long-distance feature relationship. We also propose a lightweight skip connection architecture called Residual-in-Residual Sparse Block~(RRSB) to further improve the super-resolution performance and save computation. Extensive experiments show that SPSAGAN can generate more realistic and visually pleasing textures compared to state-of-the-art SFTGAN~\cite{wang2018sftgan} and ESRGAN~\cite{wang2018esrgan} on many SR datasets.

\end{abstract}

\section{Introduction}
Single image super-resolution~(SR) is aimed to restore a high resolution~(HR) image from a single low-resolution~(LR) one. 
The problem is ill-posed because multiple solutions exist for any given LR image. Due to its superior performance, methods based on convolutional neural networks~(CNN)~\cite{Chao2014Learning,DongAccelerating,KimAccurate,Kim_2016_DRCN,LapSRN} have attracted much attention in recent years to learn the mapping from LR to HR images

To push super-resolution closer to natural images, several new losses are proposed to replace the traditional mean squared error~(MSE) which tends to encourage blurry and implausible results~\cite{Chao2014Learning,DongAccelerating,KimAccurate}. For example, the perceptual loss~\cite{Bruna2015per,Johnson2016Perceptual} has been proposed to optimize the network in a feature space instead of pixel space. Generative Adversarial Network~(GAN) loss~\cite{Christian2017Photo,enhancenet} is introduced to encourage perceptually-rich textures and significantly improves the visual quality compared with PSNR-oriented methods~\cite{Chao2014Learning,Christian2017Photo,dai2019SAN}.
\begin{figure}[ht]
	\centering
	\setlength{\abovecaptionskip}{0.1cm}
	\includegraphics[width=10.5cm]{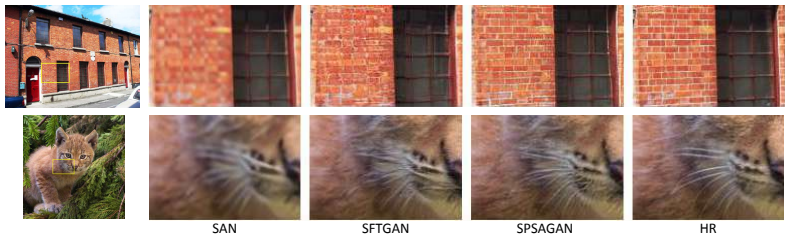}
	\caption{The SR images generated by SAN~\cite{dai2019SAN}, SFTGAN~\cite{wang2018sftgan} and SPSAGAN respectively. (Zoom in for best view).} 
	\label{fig:sft}
	\vspace{-0.3cm}
\end{figure}

One disadvantage of the above methods is the lack of prior knowledge to guide the SR algorithm. Perceptual and adversarial losses (without prior) add textures which are learned from images belonging to different categories, neglecting the semantic implications contained in the same category. 
Wang~\emph{et al.}~\cite{wang2018sftgan} address this problem and propose the Spatial Feature Transform Generative Adversarial Network~(SFTGAN) which is conditioned on segmentation probability maps to improve super-resolution. However, the receptive field of the network based on GAN structure is relatively small. Since the SR images are generated by adjacent image patches, it fails to capture semantic information from faraway patches. Figure~\ref{fig:sft} shows that SFTGAN is not satisfactory in recovering texture details, especially when textures scatter over a wide spacial range. 
Several researches aim at enlarging the receptive field of super-resolution by attention-based modules. For example, Dai~\emph{et al.}~\cite{dai2019SAN} propose a trainable second-order channel attention network~(SAN) to adaptively rescale the channel-wise features in SR. Pathak~\emph{et al.}~\cite{PathakEfficient} directly add a self-attention layer to SRGAN~\cite{Christian2017Photo} with the motivation to generate perception-friendly textures in a wide spatial range.  

\begin{figure}
	\vspace{-0.5cm}
	\centering
	\setlength{\abovecaptionskip}{0.cm}
	\includegraphics[width=12cm]{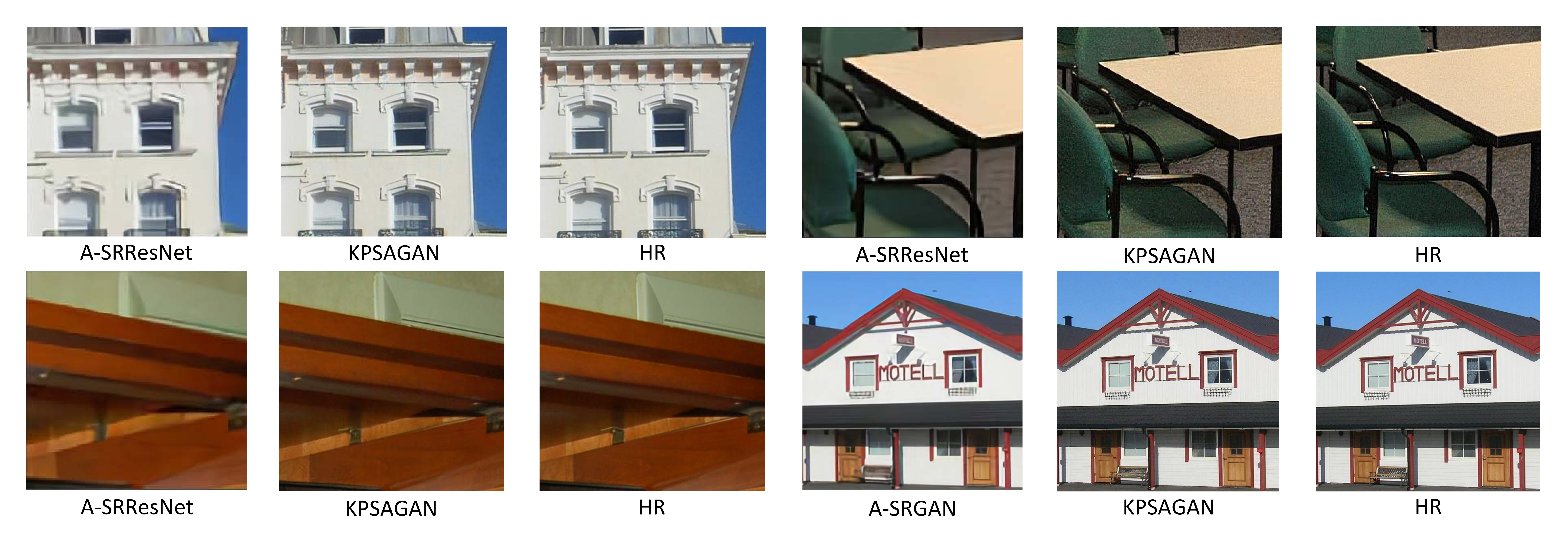}
	\caption{The SR images generated by A-SRResNet, A-SRGAN and SPSAGAN 
(Zoom in for best view).}
	\label{fig:asrgan}
	\vspace{-0.2cm}
\end{figure}

Inspired by these works, we propose Segmentation-Prior Self-Attention Generative Adversarial Network (SPSAGAN) to combine segmentation-priors and feature attentions in a unified GAN-based network. The feature attention module captures long-range and multi-level dependencies across the whole image regions, and the segmentation-prior forces the GAN generator focus on the correct segmentation categories, avoiding random generation over a large scale of image patches. The final attention maps are obtained by a carefully-designed weighted addition of segmentation and feature attentions, where weights are assigned according to the different relationship of segmentation and feature attentions. Figure~\ref{fig:sft} and \ref{fig:asrgan} show that the proposed SPSAGAN achieves better SR results than methods which only consider segmentation-prior~(SFTGAN~\cite{wang2018sftgan}) or attention (SAN~\cite{dai2019SAN}, A-SRResNeT/A-SRGAN~\cite{PathakEfficient}) respectively.

Complementary to combining segmentation-priors and self-attentions, we also investigate the network architecture design, which is another important factor for performance. The baseline of the proposed method is the Residual-in-Residual Dense Block~(RRDB)~\cite{wang2018esrgan}, which implements dense connections to reuse features of convolution layers.  Since recent researches show that unnecessary connections may affect the performance~\cite{CondenseNet}, we further investigate pruning methods to automatically eliminate unnecessary connections and make the residual blocks compact. In this paper, we introduce a lightweight skip connection structure called Residual-in-Residual Sparse Block~(RRSB) to prune unnecessary connections of RRDB. Experiments show that the pruning method boosts the performance and saves computation simultaneously.

Our contributions are three-fold:
\begin{enumerate}
  \item We propose a novel SR algorithm to combine segmentation-priors and feature attentions in a unified GAN-based network. The segmentation probability maps are combined with the self-attention mechanism by weighted addition, so that the GAN generator can emphasize textures in the same segmentation category and focus on the long-distance feature relationship.
  \item  We propose a lighter skip connection structure RRSB which prunes the dense connections of RRDB to improve performance and save computation.
  \item Extensive experiments show the effectiveness of the proposed method on different datasets compared with state-of-the-arts.

\end{enumerate}

\section{Related Work}
{\bf Single Image Super Resolution.} Convolutional neural network for super-resolution is originated from Dong~\emph{et al.}'s work SRCNN~\cite{Chao2014Learning},  
and later on various network architectures are proposed to map between low- and high-resolution images in an end-to-end manner. Dong~\emph{et al.}~\cite{DongAccelerating} propose a faster network structure FSRCNN to accelerate SRCNN. Kim~\emph{et al.}~\cite{KimAccurate} 
introduce residual learning to ease the training difficulty, which achieves significant improvement in accuracy. LapSRN~\cite{LapSRN} implements the Laplacian pyramid structure to progressively reconstruct the sub-band residuals of high-resolution images. Ledig~\emph{et al.}~\cite{Christian2017Photo} propose ResNet~\cite{Kaiming2016Deep} to construct a deeper network SRResNet. They also propose SRGAN with perceptual and GAN losses~\cite{Johnson2016Perceptual,Christian2017Photo}. 
EnhanceNet~\cite{enhancenet} further expands the SRGAN by combining automated texture synthesis and perceptual loss. ESRGAN~\cite{wang2018esrgan} enhances the SRGAN by introducing the Residual-in-Residual Dense Block~(RRDB) without batch normalization, restoring more accurate brightness and realistic textures of the SR images. By defining the naturalness prior in the low-level domain and constraining the output image in the natural manifold, NatSR~\cite{Soh_2019_CVPR_NatSR} generates more natural and realistic images compared with state-of-the-arts. Our work is an extension of the baseline network ESRGAN~\cite{wang2018esrgan} to add a segmentation-prior self-attention~(SPSA) module and make the network focus on the textures in the same segmentation category of the image. 

\noindent {\bf Attention.} The attention mechanism is widely used in image classification~\cite{WangResidual,HuSqueeze}, segmentation~\cite{OktayAttention,fu2018dual,li2019expectationmaximization}, and super-resolution~\cite{LiuAn,zhang2018rcan,dai2019SAN,Li2019,liu2019image}. Self-attention GAN~(SAGAN)~\cite{Zhang2018Self} is firstly proposed to generate images with consistent objects/scenarios for image generation tasks. 
Pathak~\emph{et al.}~\cite{PathakEfficient} introduce a flexible self-attention layer to process large-scale image super-resolution. Our method differs from pervious works in two aspects -- first, we introduce segmentation-priors to constrain the feature attention mechanism; and second, we propose a novel fusion algorithm to combine the feature and segmentation attentions with weighted addition.

\noindent {\bf Semantic Guidance.} 
Semantic information is increasingly used in various image processing tasks such as sytle transfer~\cite{gatys2017controlling}, video debluring~\cite{Zhu2017Be} and image generation~\cite{isola2017imagetoimage}. Wang~\emph{et al.}~\cite{wang2018sftgan} introduce semantic probability maps to apply conditional normalization, guiding texture recovery for different regions in the super-resolution domain.
Similarly, Wu~\emph{et al.}~\cite{semantic2019} propose semantic-prior for video super-resolution. The main difference of our method is that we enhance the guidance of segmentation network by directly using semantic probability maps to constrain feature attention, thus allowing the network focus on the same segmentation category of the reconstructed pixels.

\noindent {\bf Network Redundancy.} Many previous researches~\cite{DenilPredicting,Chen2015Compressing,HuangDeep} indicate neural networks are typically over-parameterized. Zoph~\emph{et al.}~\cite{nas} utilize reinforcement learning to find compact network structures in the search space. They also prove that complex network structure does not always result in good performance. 
The DenseNet architecture~\cite{huang2017densely} alleviates the need for feature replication by directly connecting each layer with its previous layers.  The CondenseNet~\cite{CondenseNet} simplifies DenseNet by  pruning its connections which have smaller filter importance values. The performance increases after pruning, indicating much redundancy exists in the unpruned DenseNet. 
The proposed RRSB is also based on pruning redundant connections of RRDB. However, we design a dissimilarity measure among interconnected layers to guide pruning, which is different from the filter importance measure adopted by CondenseNet. 

\section{The Proposed Method}
\begin{figure*}
	\vspace{-0.6cm}
	\setlength{\abovecaptionskip}{0.cm}
	\centering
	\includegraphics[width=12cm]{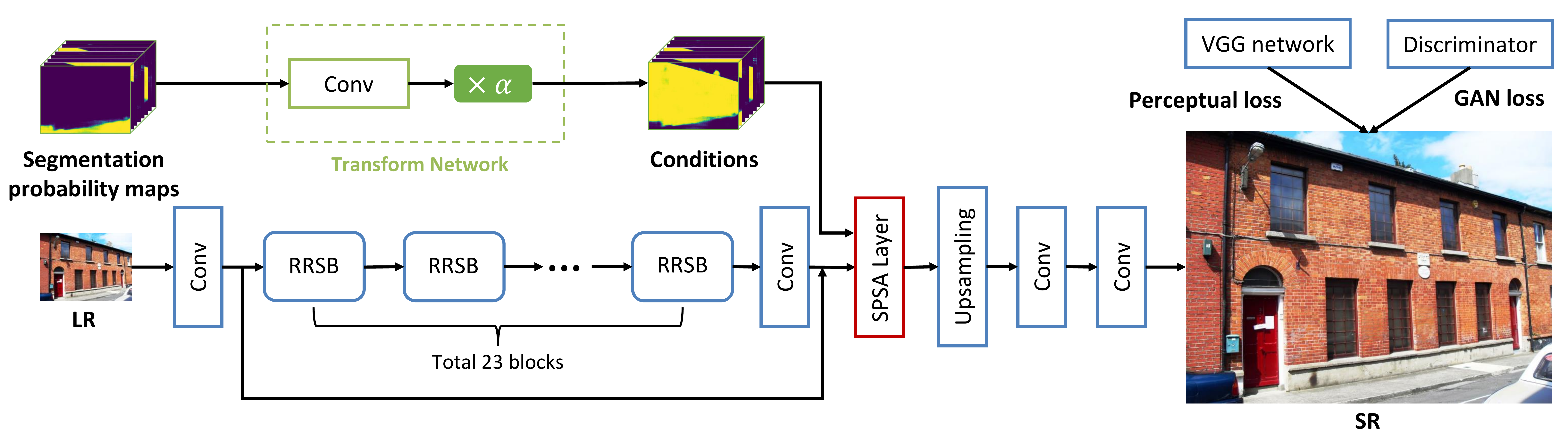}
	\caption{The structure of the proposed network. The segmentation probability maps are first fed into a transform network to convert into the same shape as the feature maps. Both of them are fed into the SPSA layer to extract attention maps. The attention maps then pass through several upsampling and convolution layers to obtain the final SR image. The perceptual and GAN losses are implemented to train the network.}
	\label{fig:structure}
	\vspace{-0.6cm}
\end{figure*}

Figure~\ref{fig:structure} shows our network architecture, which is an extension of SRGAN~\cite{Christian2017Photo} and ESRGAN~\cite{wang2018esrgan}. The LR images are fed into a CNN with~$23$ basic blocks to obtain feature maps. In SRGAN~\cite{Christian2017Photo}, the basic blocks are plain convolution layers. ESRGAN~\cite{wang2018esrgan} updates the basic block with residual-in-residual dense block~(RRDB), which combines multi-level residual network and dense connections to improve performance. For the proposed method, we replace RRDB with a lighter skip connection structure RRSB to further improve performance. The obtained feature maps after basic blocks are combined with segmentation-priors in the proposed SPSA layer to balance the influence of segmentation and feature attentions. The output of the SPSA layer is passed to the upsampling layer and then several convolution layers to obtain the SR image. Inspired by~\cite{wang2018esrgan,Christian2017Photo}, we apply the perceptual loss~\cite{wang2018esrgan} generated by a pretrained VGG-19 network~\cite{SimonyanVGG}. The GAN loss is also added to the network to make reconstruction more natural. 
Following~\cite{Christian2017Photo,enhancenet}, we apply a VGG-style~\cite{SimonyanVGG} network with Leaky ReLU
activations for the discriminator of GAN. The novelty of the proposed architecture is the SPSA layer and the RRSB which prunes the dense RRDB to improve performance and save computation. 


\subsection{Segmentation-prior Self-Attention~(SPSA)}
\begin{figure}[ht]
	\vspace{-0.6cm}
    \setlength{\abovecaptionskip}{0.1cm}
	\centering
	\includegraphics[width=8cm]{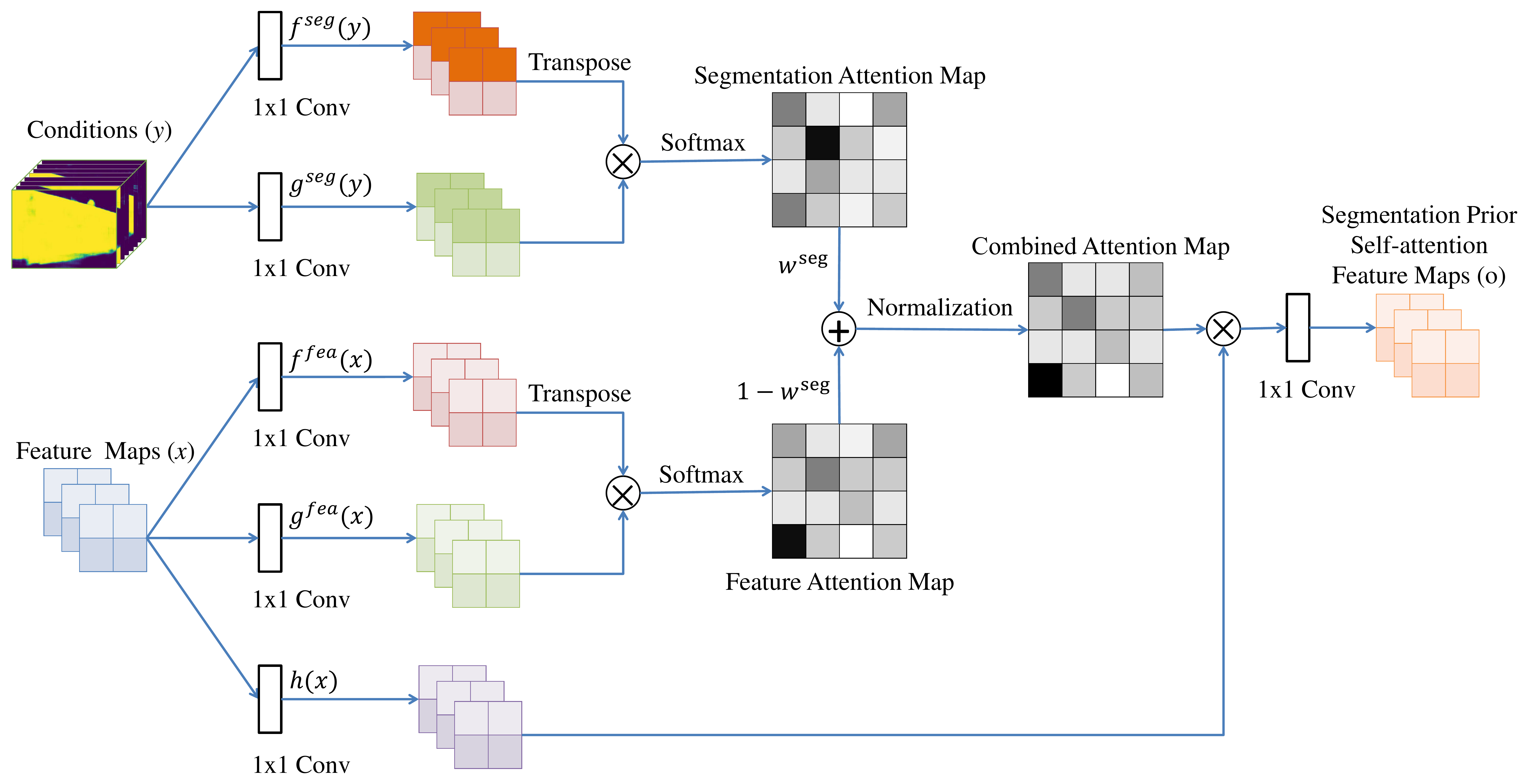}
	\caption{The proposed segmentation-piror self-attention module for the SPSAGAN.} 
	\label{fig:attention}
	\vspace{-0.2cm}
\end{figure}
\label{sec:seg}
Figure~\ref{fig:attention} shows the structure of the proposed SPSA module. The feature map from the previous layer is represented as~$\bm{x} \in \mathbb{R}^{C \times N}$, where~$C$ and~$N$ represent number of channels and number of pixels respectively. It is first transformed into two feature spaces~$\bm{f}$ and~$\bm{g}$ by~$1\times1$ convolutions:
\begin{equation}
\bm{f}^{fea}(\bm{x}) = {\bm{W_f}}^{fea}\bm{x} \textrm{, }\bm{g}^{fea}(\bm{x}) = {\bm{W_g}}^{fea}\bm{x}.
\end{equation}
The feature attention~$\beta^{fea}$ between the~$j$th and~$i$th pixels is calculated as:
\begin{equation}
\label{eqn:fea attention}
\beta_{j,i}^{fea}= \frac{exp(s_{ij}^{fea})}{ \sum_{i=1}^N exp(s_{ij}^{fea})}, \textrm{ where } s_{ij}^{fea} = \bm{f}^{fea}(\bm{x_i})^T\bm{g}^{fea}(\bm{x_j}).
\end{equation}

For calculating the segmentation-prior knowledge, the LR image is interpolated to the size of HR image by bicubic kernels, then fed into a semantic segmentation network~\cite{LiuSeg} which is pretrained on COCO dataset~\cite{Lin2014Microsoft} and fine-tuned on ADE dataset~\cite{Zhou2017Scene}. 
The network is trained to segment outdoor scenes with seven categories: sky, mountain, plant, grass, water, animal and building. Pixels that fall outside of the seven categories are labeled as ‘background’.

The segmentation probability map~$\bm{m_{seg}}$ is first sent into a transform network which consists of a convolution layer and a scale layer to obtain segmentation features:
\begin{equation}
\bm{z} = \alpha Conv(\bm{m_{seg}}).
\end{equation}
The number of filters is~$C$, and the size of each filter is~$4\times4$ with stride $4$ to ensure the dimension of~$\bm{z}$ is~$C\times N$, the same with the feature map~$\bm{x}$. Here~$\alpha$ is a trainable parameter to make the magnitude of~$\bm{z}$ comparable to~$\bm{x}$ to guarantee fast convergence. The initial settings of~$\alpha$ is the average~$L_1$ norm of~$\bm{x}$ divided by the average~$L_1$ norm of~$Conv(\bm{m_{seg}})$.     
Similar with the feature attention, $\bm{z}$ is first transformed into two feature spaces to calculate the segmentation attention $\beta^{seg}$, where~$\bm{f}^{seg}(\bm{z}) = {\bm{W_f}}^{seg}\bm{z}$ and~$\bm{g}^{seg}(\bm{z}) = {\bm{W_g}}^{seg}\bm{z}$.
\begin{equation}
\label{eqn:seg attention}
\beta_{j,i}^{seg}= \frac{exp(s_{ij}^{seg})}{ \sum_{i=1}^N exp(s_{ij}^{seg})},\textrm{ where }s_{ij}^{seg} = \bm{f}^{seg}(\bm{z_i})^T\bm{g}^{seg}(\bm{z_j})
\end{equation}

The feature and segmentation attention maps are combined by the weighted sum rule, where weights are automatically calculated by
\begin{equation}
\label{eqn:weight}
w^{seg}_{j,i} = \frac{\left| \beta^{seg}_{j,i}-\beta^{fea}_{j,i} \right| }{\beta^{seg}_{j,i}+\beta^{fea}_{j,i}}, \textrm{  } w^{fea}_{j,i} = 1-w^{seg}_{j,i}.
\end{equation}
The combined attention is obtained and normalized by
\begin{equation}
\label{eqn:attention}
\beta_{j,i} = w^{seg}_{j,i} \beta^{seg}_{j,i} + (1-w^{seg}_{j,i}) \beta^{fea}_{j,i}, \textrm{  } \beta_{j,i}= \frac{\beta_{j,i}}{ \sum_{i=1}^N \beta_{j,i}}.
\end{equation}
The reason for assigning~$w^{seg}_{j, i}$ by Equation~(\ref{eqn:weight}) lies in four aspects: (1) When~$\beta^{seg}$ and $\beta^{fea}$ are relatively similar, the guidance of segmentation-prior is neglected because feature attention is consistent with the segmentation attention.  In this situation, $w^{fea}_{j,i}$ should be increased to enhance the influence of~$\beta^{fea}$ with the motivation that feature attentions are helpful in generating texture details of the SR image. 
(2) When~$\beta^{seg}$ is smaller and $\beta^{fea}$ bigger, it means that colors or textures of two regions are similar, but they belong to different categories. In this situation, the guidance of the segmentation-prior should take effect to de-emphasize the interference of different segmentation categories. (3) The situation rarely happens when~$\beta^{seg}$ is bigger and $\beta^{fea}$ smaller. For a single image, features from the same segmentation category are likely to be similar. Even if it happens, emphasizing the segmentation attention is also a good solution because pixels belonging to the same category tends to complement with each other. (4) The value of~$w^{seg}$ is in the range of~$[0,1]$, which is a mandate for weighted sum combination. 


Finally, the output of the SPSA layer $o = (\bm{o_1},\bm{o_2},...,\bm{o_j},...,\bm{o_N}) \in \mathbb{R}^{C \times N}$ is obtained by
\begin{equation}
\label{eqn:out fea}
\bm{o_j} = \sum_{i=1}^N \beta_{j,i}\bm{h}(\bm{x_i}),
\end{equation}
where~$\bm{h}(\bm{x_i}) = \bm{W_hx_i}$, which is also a~$1\times 1$ convolution of feature map~$\bm{x}$.

\subsection{The Design of Residual-in-Residual Sparse Block}
\label{sec:network arch}
\begin{figure}[ht]
	\vspace{-0.5cm}
	\setlength{\abovecaptionskip}{0.1cm}
	\centering
	\includegraphics[width=11cm]{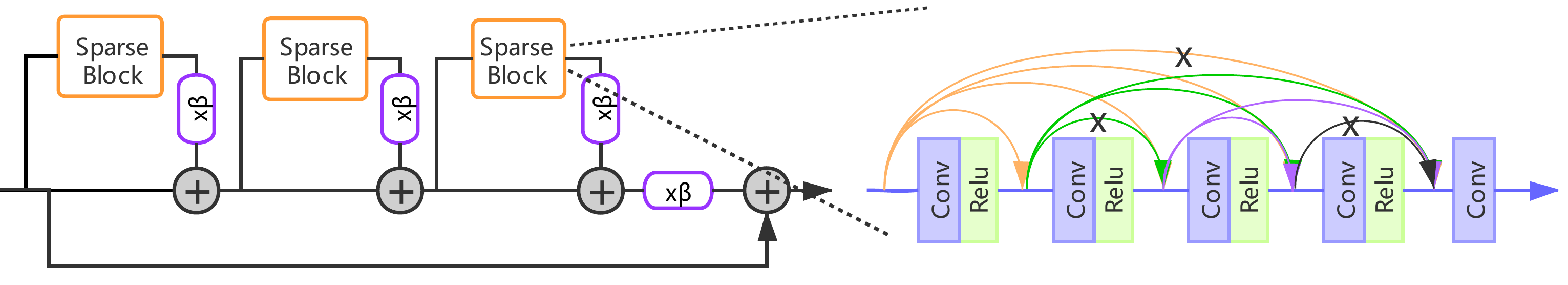}
	\caption{Sparse block is used in RRSB module. The ‘x' means that the connection is pruned from the original RRDB. Here $\beta$ is the residual scaling parameter of RRDB.}
	\label{fig:rrsb}
	\vspace{-0.2cm}
\end{figure}

The proposed residual-in-residual sparse block (RRSB) is originated from the residual-in-residual dense block~(RRDB)~\cite{wang2018esrgan}. As shown in Fig.~\ref{fig:rrsb}, each RRDB consists of three dense blocks and each dense block consists of five convolution layers with dense connections for each layer. The dense connections consume much computation and may be redundant. In this paper, we propose the RRSB which aims at pruning redundant connections in RRDB. 

For a dense block consisting of~$K$ convolution layers, we denote its input as~$x_0$, and the subsequent output of the~$l$th layer 
as~$x_l$. In RRDB, the $l$th layer receives feature-maps from all preceding layers as input:
$x_l = H_l([x_0, x_1, ..., x_{l-1}])$, where $[x_0, x_1, ..., x_{l-1}]$ refers to the concatenation of feature-maps.  Here~$H_l(·)$ is a composite function consisting of a $3\times3$ convolution and leaky ReLU. 
The dissimilarity measure of the feature maps~$x_i$ ($i=1 \sim l-1$) and~$x_l$ is defined as
\begin{equation}
\label{eqn:DS}
DS_{i-l} =  \frac{\left\| x_i - x_l \right\|_2}{ \sum_{p=1}^{l-1} \left\| x_p - x_l  \right\|_2}.
\end{equation}
For the~$l$th layer, the associated connections to be pruned are generally those with smaller dissimilarity measures. If~$x_i$ is similar with $x_l$, it is unnecessary to concatenate~$x_i$ to $x_l$. Due to the different number of preceding connections, it is difficult to set a fixed threshold. Thus, we use a heuristic method to determine the threshold. Firstly, the~$l-1$ dissimilarity measures are clustered into two classes by K-means algorithm, then all connections in the class with smaller mean are removed from the network. One exception is that the dissimilarity difference between two classes is not prominent, i.e., the difference is less than~$5\%$. In this situation, all connections are retained without pruning.

The network is firstly trained with RRDB till convergence. Then, the average dissimilarity measures during last several iterations are implemented to prune connections and obtain the basic RRSB blocks. The final network with RRSB is trained from scratch till convergence.

\section{Experiments}
\label{sec:exp}
For preprocessing, the spatial size of the HR images are randomly cropped with size~$96\times96$ from training datasets, and then down-sampled with a scaling factor of~$4$ to obtain~$24\times24$ LR images. 
The training process is divided into two steps. Firstly, we pre-train a PSNR-oriented model using the~$L_1$ loss without the SPSA module. 
The learning rate is initialized as $2\times10^{-4}$ and decayed by a factor of 2 every $2\times10^{5}$ iterations.  
The pre-trained model is employed as the initialization for the proposed method. For further training of SPSAGAN, we use Adam~\cite{kingma2014adam} with $\beta_1 = 0.9$. The learning rate is set to~$5\times10^{-4}$ for the self-attention module and~$1\times10^{-4}$ for the rest of the network. 
The learning rate decays by a factor of $2$ every $100$k iterations. The batch size is set to~$16$ consistently for the two steps. 


We use the DIV2K~\cite{Agustsson2017NTIRE} and Flickr2K~\cite{Flickr} datasets for pre-training, which contain~$800$ and~$2,650$ $2$K resolution images respectively. Then, we use the OST training set~\cite{wang2018sftgan} to train the proposed SPSAGAN. OST contains outdoor scenes with seven categories, which are the same with the training dataset of the segmentation network~\cite{LiuSeg}. Each category has $1$k to $2$k images and the total image number is $10,324$. One disadvantage of OST is that each image only contains one category, so it is impossible to learn category relationship in a single image. However, from Equation~(\ref{eqn:seg attention}), category relationship is important for the proposed attention-based method. To remedy this, we randomly select training images from DIV2K which contain multiple categories for SPSAGAN training. Following~\cite{wang2018sftgan}, the ratio of OST and DIV2K data samples are set to~$10:1$. 

\subsection{The Self-Attention Mechanism}
\begin{figure}[ht]
	\centering
	\includegraphics[width=12cm]{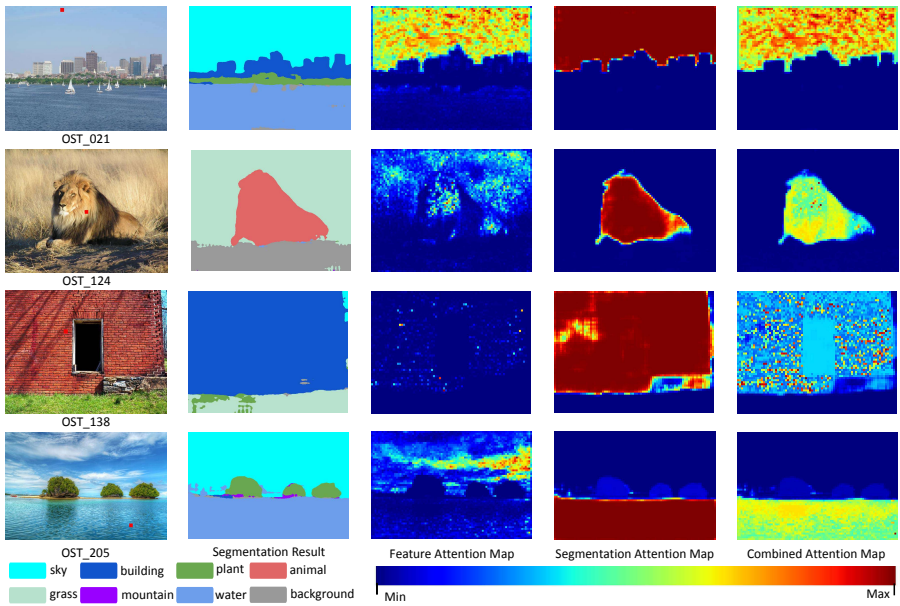}
	\caption{The image, segmentation result and attention maps of OST dataset.}
	\label{fig:attmap}
\end{figure}

Figure~\ref{fig:attmap} is the visualization of attention maps. The red point on the image is the query pixel, e.g., the~$j$th pixel in Equation~(\ref{eqn:attention}). The output of segmentation results is calculated as the maximum of the eight segmentation probability maps. The feature, segmentation and combined attention maps are the~$\beta$'s in Equation~(\ref{eqn:fea attention}), (\ref{eqn:seg attention}) and~(\ref{eqn:attention}) respectively, where the~$i$th pixel goes through the whole image. 
It shows that the attention maps tend to concentrate on long-range pixels rather than spatially local pixels. For example, in line 1, the combined attention focuses on the whole sky; and in line 2, the query point attends to the grass and lion in the feature attention map. 
These long-range dependencies cannot be captured by convolution with local receptive fields. We also find that the feature attention module tends to concentrate on similar color and texture regions, but the segmentation attention tends to inference according to categories. The fourth line illustrates one example that the segmentation attention guides the feature attention to focus on textures of the same category. 
The query pixel locates in the water region, but the feature attention is misled by part of the sky region because their colors are similar. However, the segmentation  attention takes effect to pull the combined attention back into the water region. The second line also shows that even if the feature attention is randomized over the image, the combined attention can still be reasonable due to the interference of the segmentation attention. 
These observations further demonstrate that the segmentation-prior is complementary to feature map convolution, which can bring robustness to super-resolution. 
The attention maps of the out-of-category images are provided in the supplementary material.

\subsection{Comparison with the State-of-the-art}
The proposed SPSAGAN is compared with several PSNR-oriented methods including SRCNN~\cite{Chao2014Learning}, SRResNet~\cite{Christian2017Photo}, SAN~\cite{dai2019SAN}, and also with several perception-driven approaches including SRGAN~\cite{Christian2017Photo}, NatSR~\cite{Soh_2019_CVPR_NatSR}, SFTGAN~\cite{wang2018sftgan} and ESRGAN~\cite{wang2018esrgan}. The datasets for comparison are OST, Set5, Set14 and BSD100.

Three quantitative metrics are implemented for evaluation, i.e., PSNR~(dB), SSIM~\cite{Wang2004Image} (evaluated on the Y channel in YCbCr color space) and the Perceptual Index (PI)~\cite{PRIM} (lower PI stands for better perceptual quality). Table~\ref{qr} summarizes the average of these metrics for each method. PSNR-oriented approaches yield better PSNR and SSIM values, but perception-driven methods achieve better PI values.  The proposed SPSAGAN achieves the best PI on BSD100, and the PI on OST is also close to the best method SFTGAN. However, PI is not always the superior metric for super-resolution. 
For example, it is unreasonable that the PI of SFTGAN on OST is even better than the ground truth HR. The above experiments indicate that designing a unanimously agreed quantitative metric for super-resolution is still an unsolved problem.
\begin{table}[ht]
	\centering
	\caption{Quantitative evaluation of PSNR, SSIM and PI on BSD100 and OST. The best and second best results are {\bf highlighted} and \underline{underlined}, respectively. [$4\times$ upscaling]}
	\label{qr}
	\scriptsize
	\begin{tabular}{lllllllllll}
		{\bf BSD100} & Bicubic & SRCNN & SRResNet & SAN & SRGAN & NatSR & SFTGAN & ESRGAN & SPSAGAN & HR \\
		\midrule
		{\bf PSNR} & $25.96$ & $26.90$ & $27.58$ & $\mathbf{27.78}$ & $25.16$ & $26.44$ & $25.33$ & $\underline{27.76}$ & $25.16$ & $\infty$ \\
		{\bf SSIM} & $0.6675$ & $0.7101$ & $\mathbf{0.7620}$ & $\underline{0.7436}$ & $0.6688$ & $0.6827$ & $0.6510$ & $0.7432$ & $0.6344$ & $1.000$ \\
		{\bf PI} & $7.003$ & $6.045$ & $5.437$ & $5.239$ & $2.565$ & $2.780$ & $\underline{2.396}$ & $2.510$ & $\mathbf{2.365}$ & $2.310$ \\
	\end{tabular}
	\\[0.4cm]
	\begin{tabular}{lllllllllll}
		{\bf OST} & Bicubic & SRCNN & SRResNet & SAN & SRGAN & NatSR & SFTGAN & ESRGAN & SPSAGAN & HR \\
		\midrule
		{\bf PSNR} & $25.75$ & $24.50$ & $\underline{26.78}$ & $\mathbf{27.23}$ & $24.23$ & $25.86$ & $24.71$ & $24.78$ & $24.50$ & $\infty$ \\
		{\bf SSIM} & $0.6635$ & $0.7032$ & $\underline{0.7221}$ & $\mathbf{0.7376}$ & $0.6270$ & $0.6743$ & $0.6340$ & $0.6500$ & $0.6242$ & $1.000$ \\
		{\bf PI} & $6.780$ & $6.079$ & $5.307$ & $5.525$ & $\underline{2.286}$ & $2.545$ & $\mathbf{2.276}$ & $2.550$ & $2.358$ & $2.500$ \\
	\end{tabular}
\end{table}

\begin{figure}
	\centering
	\includegraphics[width=11cm]{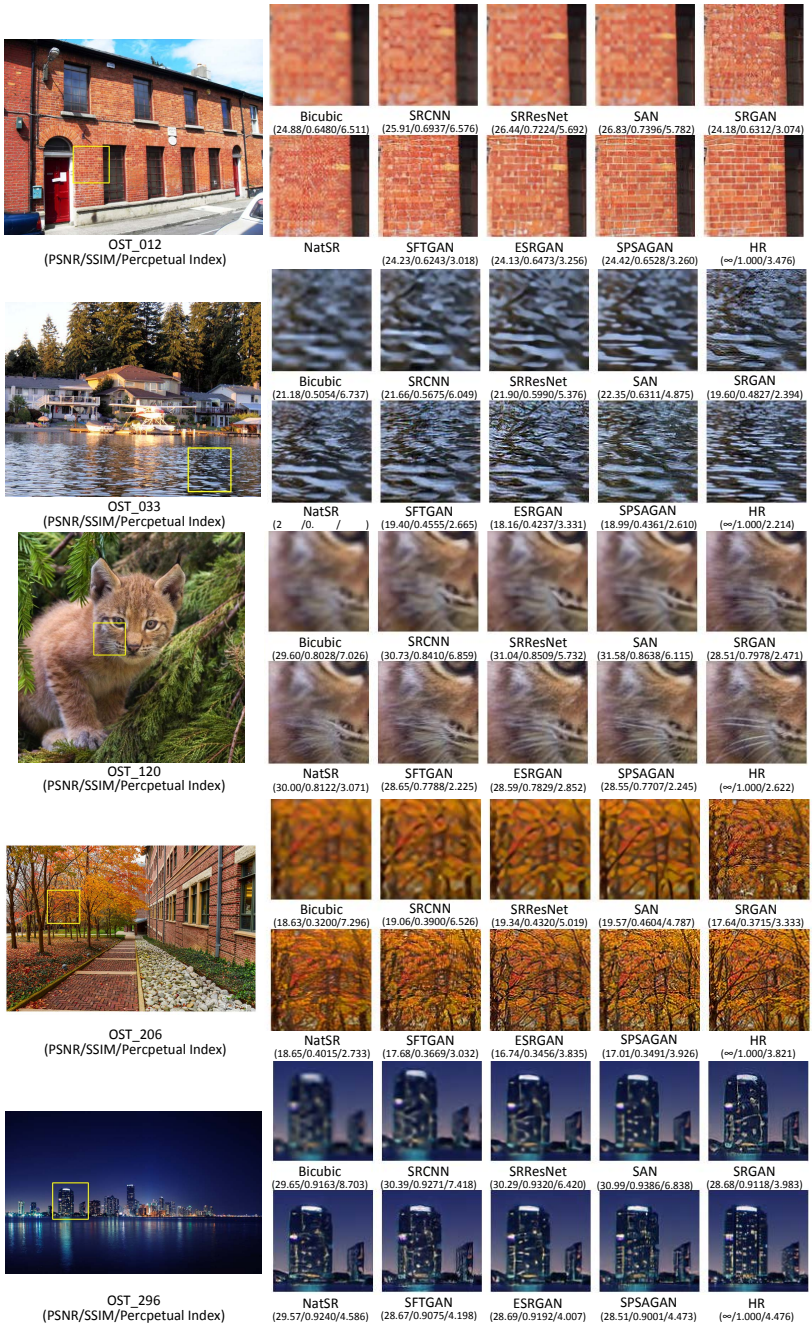}
	\caption{Qualitative results of different methods (Zoom in for best view).} 
	\label{fig:final_result}
\end{figure}

Figure~\ref{fig:final_result} shows the qualitative results of each methods.  It can be seen that the proposed SPSAGAN is superior to the previous approaches in both details and natural textures. For instance, SPSAGAN can produce more natural water waves for OST$\_033$ and more vivid textures for OST$\_206$.  
SPSAGAN is also capable of generating more detailed building structures for OST$\_012$ while other methods either produce blurry textures or the lines of bricks are not natural.

\begin{figure}[ht]
	\centering
	\includegraphics[width=11cm]{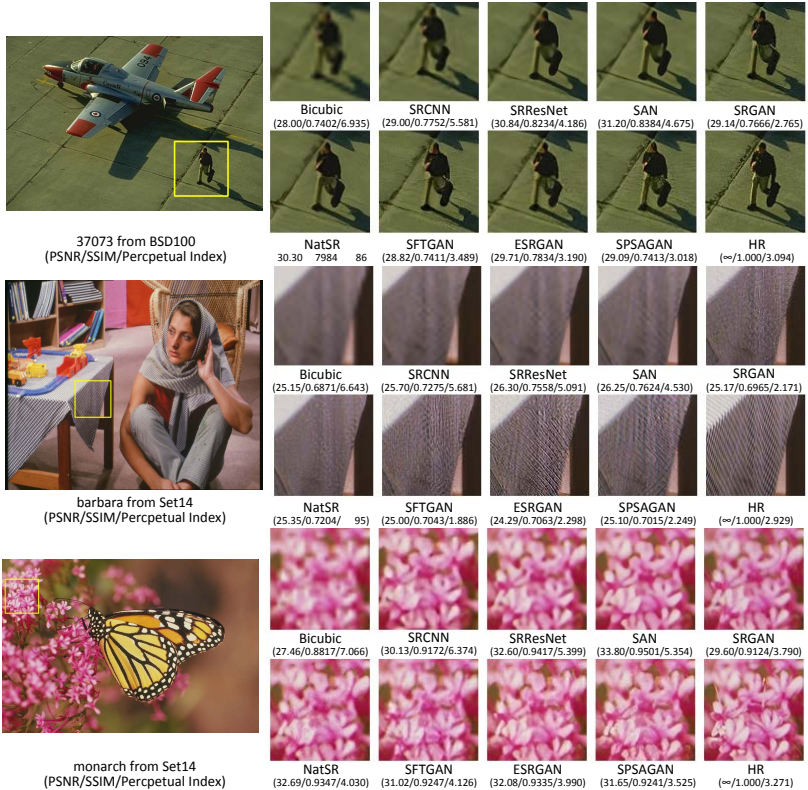}
	\caption{Qualitative results of SR methods on out-of-category images (Zoom in for best view).}
	\label{fig:re_else}
\end{figure}

Figure~\ref{fig:re_else} shows the qualitative results for image patches which are out of the seven segmentation categories (walking person, tablecloth and flower). SPSAGAN is also reliable in producing comparable results like other methods in this situation, although its performance is not as good as processing images from the seven categories. The reasonable performance is attributed to the feature attention, which still takes effects because other categories share similar textures and colors with the seven categories. However, the performance of SPSAGAN degrades because the guidance of segmentation attention is weakened in this situation. More results of out-of-category images are provided in the supplementary material.

\subsection{User Study}
We conduct the user study to compare the perceptual quality of the generated SR images. We divide the study into the following two tasks.   

\begin{figure}
	\centering
	\setlength{\abovecaptionskip}{0.cm}
	\subfigure[]{
		\begin{minipage}[c]{0.5\textwidth}
			\centering
			\includegraphics[height=3cm,width=5cm]{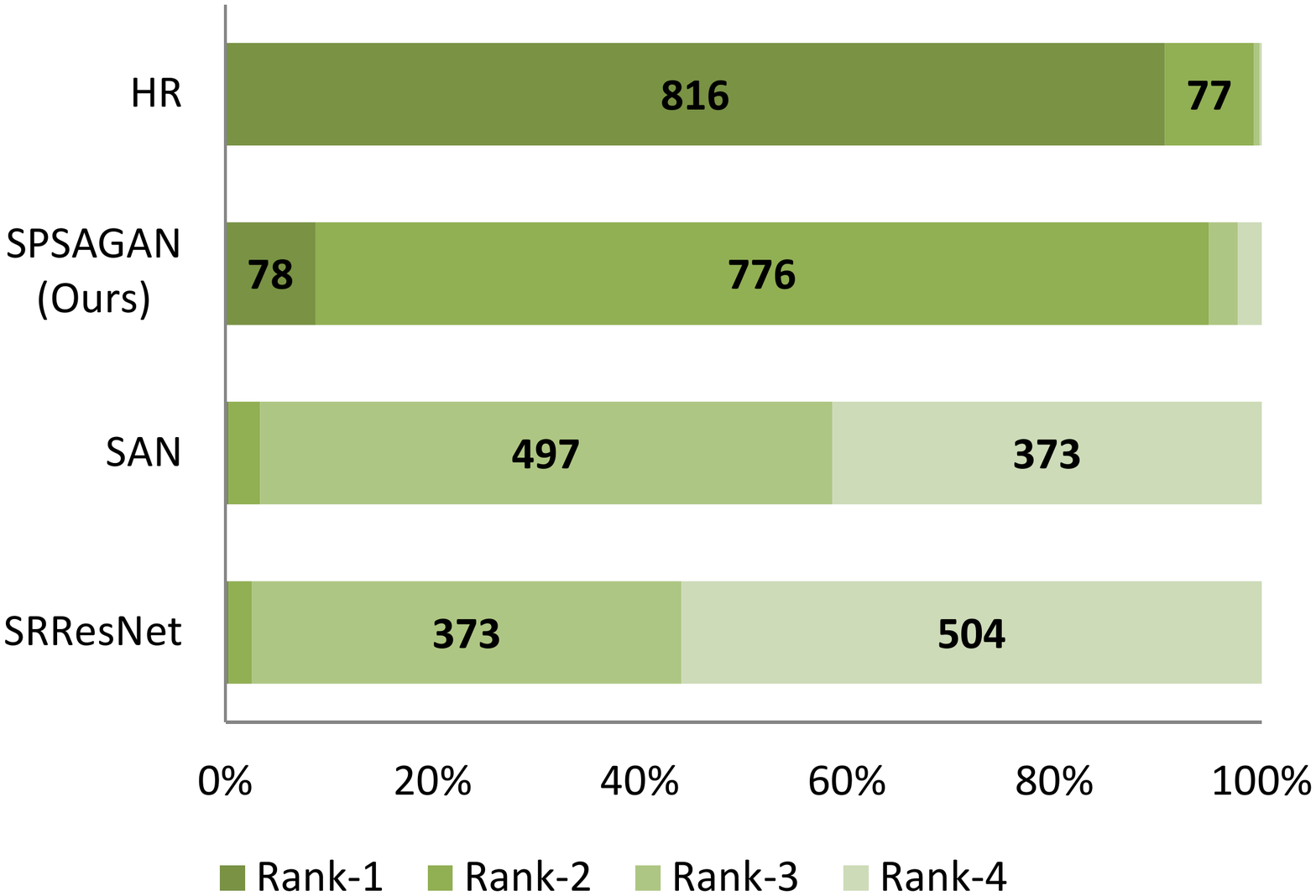}
	\end{minipage}}%
	\subfigure[]{
		\begin{minipage}[c]{0.5\textwidth}
			\centering
			\includegraphics[height=3cm,width=5cm]{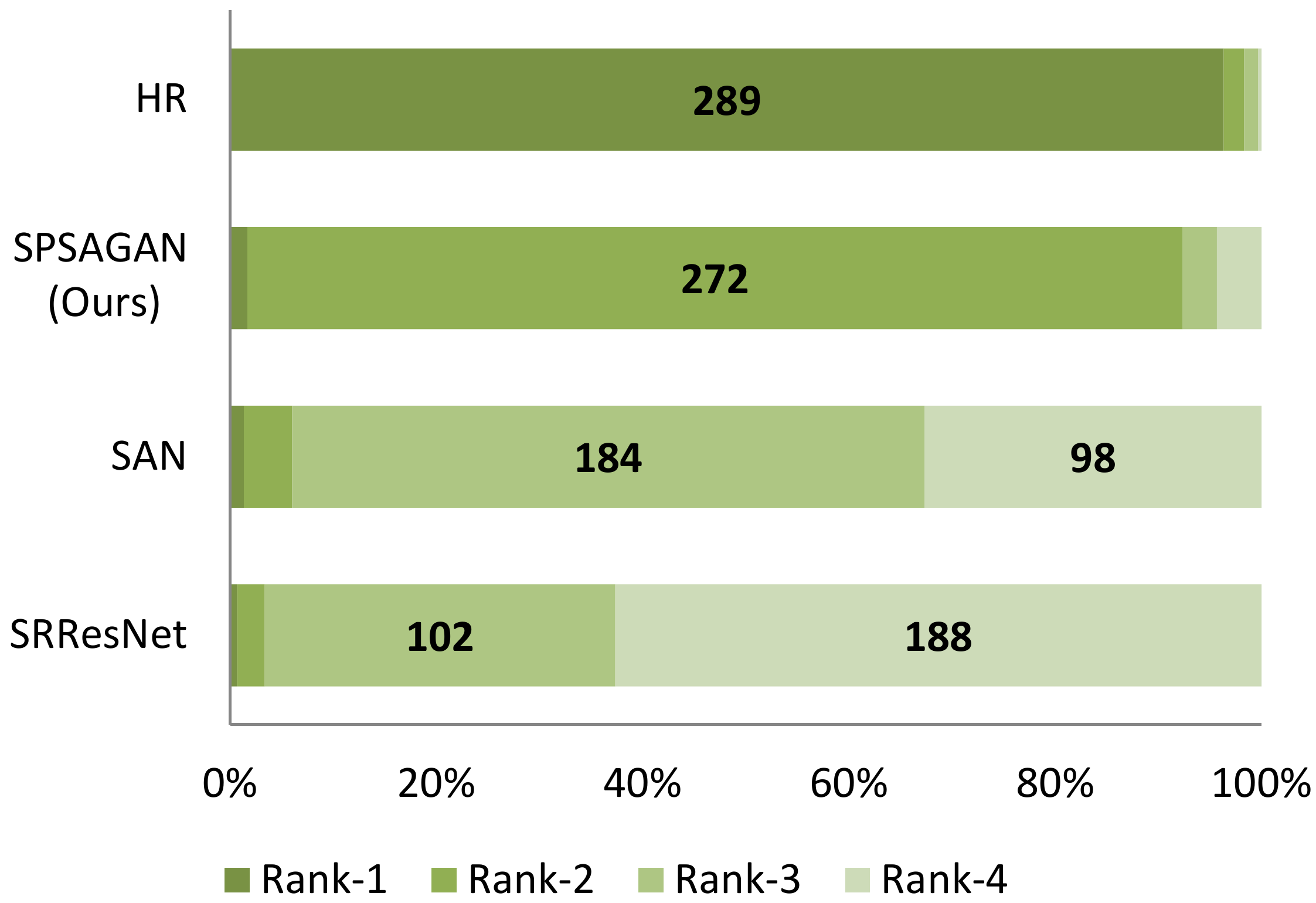}
	\end{minipage}}
	\caption{The ranking results of~SRResNet~\cite{Christian2017Photo}, SAN~\cite{dai2019SAN}, SPSAGAN and HR. Numbers represent frequency of voting. (a) The results of 30 images in OST, totally 900 valid votes. (b) The results of 10 out-of-category images, totally 300 valid votes.}
	\label{fig:uesr_psnr}
\end{figure}

Task 1 is to compare the proposed SPSAGAN with the PSNR-oriented methods. 
In this task, $30$ users are asked to rank the four images based on their visual quality -- the SR images generated by SRResNet~\cite{Christian2017Photo}, SAN~\cite{dai2019SAN} and the proposed SPSAGAN respectively, and the ground truth HR image. 
For each person, we randomly select~$30$ images from the OST test dataset and~$10$ out-of-category images from Set5, Set14 and BSD100. For each image, we show the four SR and HR images with random order to the users, and ask them to rank from~$1$ to~$4$ according to their visual quality. 
The ranking results are shown in Figure~\ref{fig:uesr_psnr}.  It can be seen that SPSAGAN is significantly better than the two PSNR-oriented methods. The only exception is that on the out-of-category images, SPSAGAN has slightly less Rank 1 images compared with SRResNet, but Rank 2 images of SPSAGAN are much more than the other two.  
On the OST dataset, sometimes SPSAGAN can confuse the users and make them think it is better than the ground truth.

\begin{figure}
	\centering
	\setlength{\abovecaptionskip}{0.cm}
	\subfigure[]{
		\begin{minipage}[c]{0.5\textwidth}
			\centering
			\includegraphics[height=2.3cm,width=5.3cm]{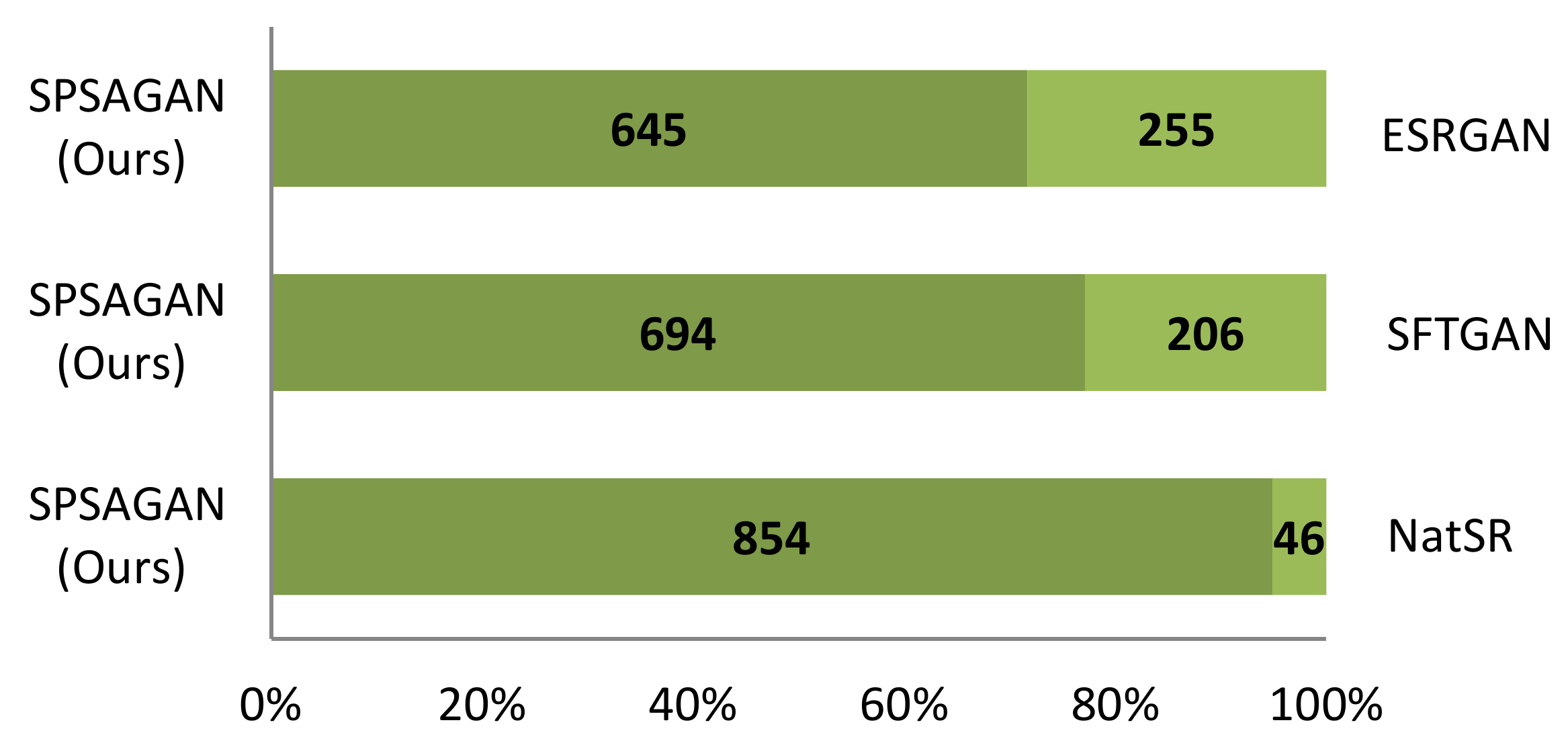}
	\end{minipage}}%
	\subfigure[]{
		\begin{minipage}[c]{0.5\textwidth}
			\centering
			\includegraphics[height=2.3cm,width=5.3cm]{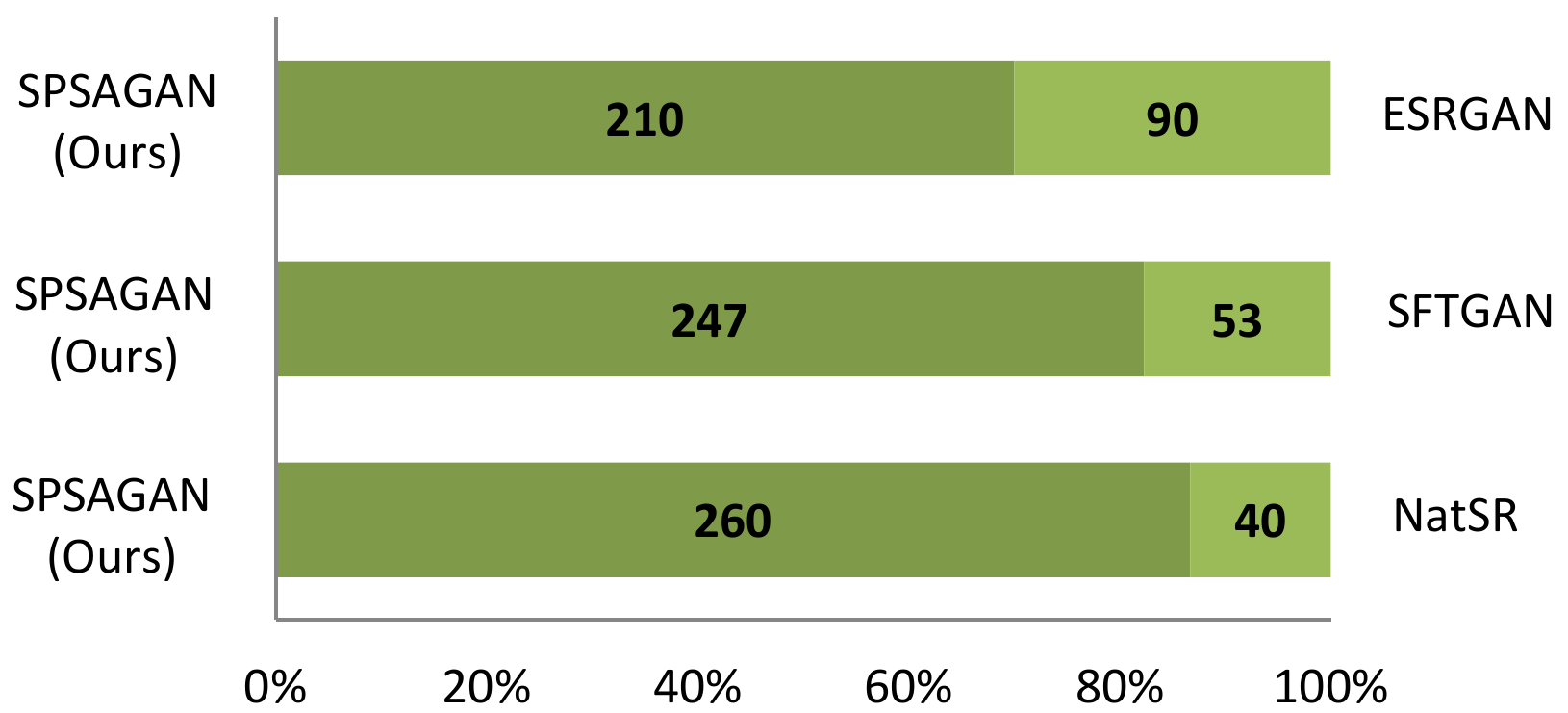}
	\end{minipage}}
	\caption{The comparison results of SPSAGAN with SFTGAN, ESRGAN and NatSR. Numbers represent the frequency of voting. (a) The results of 30 images in OST, totally 900 valid votes. (b) The results of 10 out-of-category images, totally 300 valid votes.}
	\label{fig:uesr_gan}
\end{figure}

Task 2 is to compare the generated texture quality of SPSAGAN with other perception-driven approaches. The same with Task 1, we randomly select~$30$ images from the OST test dataset and~$10$ out-of-category images from Set5, Set14 and BSD100. These images are shown by pairs, of which one is the SR image generated by the proposed SPSAGAN, and the other is generated from SFTGAN~\cite{wang2018sftgan}, ESRGAN~\cite{wang2018esrgan}, and NatSR~\cite{Soh_2019_CVPR_NatSR} respectively. We show enlarged texture patches to~$30$ users and ask them to select the image with more natural and perception-friendly textures. Figure~\ref{fig:uesr_gan} shows the comparison results. Our method ranks much higher than the other three in this situation, indicating it is superior in generating natural and visual pleasing images.

\subsection{Ablation Study}
\begin{figure}[h]
	\centering
	\includegraphics[width=12cm]{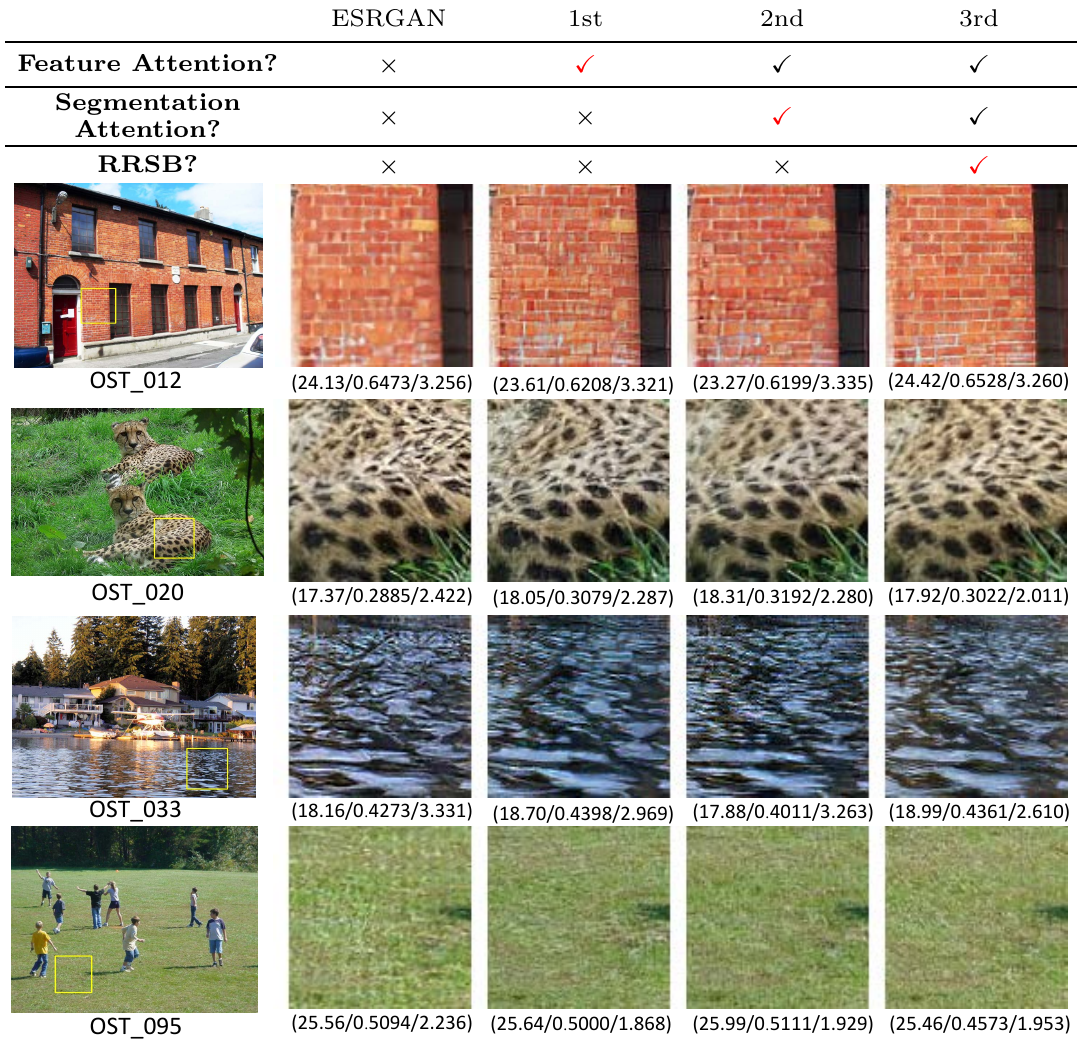}
	\caption{Visual comparison of  each component in SPSAGAN (Zoom in for best view). Each column represents a model with its configurations in the top. The red sign indicates the main improvement compared with the previous model. The numbers below the image are PSNR, SSIM and PI, respectively.}
	\label{fig:abl}
\end{figure}

To study the impact of each component in the proposed SPSAGAN, we update the baseline ESRGAN~\cite{wang2018esrgan} by gradually adding components. Figure~\ref{fig:abl} shows the visual comparison of different models. Each column represents a model with its configuration shown at the top. The red check mark indicates the major improvement compared to the previous model.

\noindent {\bf Feature Attention.} The main effect of adding feature attention is to clear the blurred texture (e.g., OST$\_012$) and eliminate strange artifacts (e.g., OST$\_020$, OST$\_033$ and OST$\_095$). The addition of feature attention expands the receptive field of the network. The generation of the current pixel is not only based on adjacent image patches, but also relies on textures from faraway patches.

\noindent {\bf Segmentation Attention.} It can be seem that segmentation attention produces clearer and more regular textures because it constrains feature attention to focus only on the textures belonging to the same segmentation category. For example, in OST$\_012$ the edges of the bricks become more flat, and the leopard's markings in OST$\_020$ are not messy.

\noindent {\bf RRSB.} Pruning unnecessary connections of RRDB further improves the overall visual quality. Some textures become soft and smooth, which is more amenable to human visual system, such as the water wave in OST$\_033$. Pruing of RRDB also saves computation. The average inference time per image before and after pruning on the OST dataset is $0.368$s and $0.296$s respectively~(tested on GeForce GTX 1080Ti). 


\section{Conclusions}
We propose a novel segmentation-prior self-attention (SPSA) layer that enables the super-resolution network to reconstruct high-quality images. The self-attention mechanism expands the receptive field of the network, and the segmentation priors constrain the focus of the attention module on regions belonging to the same segmentation category.  
We also explore the basic blocks in the network and propose a skip connection architecture to eliminate redundancy of the network, thus achieving better performance and saving computation. 
Extensive experiments demonstrate the superior performance of the proposed method in generating natural and perception-friendly SR images compared with state-of-the-arts.

%
%
\bibliographystyle{splncs04}
\bibliography{reference}

\begin{thebibliography}{10}
\providecommand{\url}[1]{\texttt{#1}}
\providecommand{\urlprefix}{URL }
\providecommand{\doi}[1]{https://doi.org/#1}

\bibitem{Agustsson2017NTIRE}
Agustsson, E., Timofte, R.: {NTIRE} 2017 challenge on single image
  super-resolution: Dataset and study. In: The IEEE Conference on Computer
  Vision and Pattern Recognition Workshops (2017)

\bibitem{PRIM}
Blau, Y., Mechrez, R., Timofte, R., Michaeli, T., Zelnik-Manor, L.: The {PIRM}
  challenge on perceptual super resolution.
  \url{https://www.pirm2018.org/PIRM-SR.html} (2018)

\bibitem{Bruna2015per}
Bruna, J., Sprechmann, P., Lecun, Y.: Super-resolution with deep convolutional
  sufficient statistics. In: International Conference on Learning
  Representations (2015)

\bibitem{Chen2015Compressing}
Chen, W., Wilson, J.T., Tyree, S., Weinberger, K.Q., Chen, Y.: Compressing
  neural networks with the hashing trick. In: International Conference on
  Machine Learning

\bibitem{dai2019SAN}
Dai, T., Cai, J., Zhang, Y., Xia, S.T., Zhang, L.: Second-order attention
  network for single image super-resolution. In: The IEEE Conference on
  Computer Vision and Pattern Recognition (2019)

\bibitem{DenilPredicting}
Denil, M., Shakibi, B., Dinh, L., Ranzato, M., De~Freitas, N.: Predicting
  parameters in deep learning. In: International Conference on Neural
  Information Processing Systems (2013)

\bibitem{Chao2014Learning}
Dong, C., Loy, C.C., He, K., Tang, X.: Learning a deep convolutional network
  for image super-resolution. In: European Conference on Computer Vision (2014)

\bibitem{DongAccelerating}
Dong, C., Loy, C.C., Tang, X.: Accelerating the super-resolution convolutional
  neural network. In: European Conference on Computer Vision (2016)

\bibitem{fu2018dual}
Fu, J., Liu, J., Tian, H., Li, Y., Bao, Y., Fang, Z., Lu, H.: Dual attention
  network for scene segmentation. In: The IEEE Conference on Computer Vision
  and Pattern Recognition (2019)

\bibitem{gatys2017controlling}
Gatys, L.A., Ecker, A.S., Bethge, M., Hertzmann, A., Shechtman, E.: Controlling
  perceptual factors in neural style transfer. In: The IEEE Conference on
  Computer Vision and Pattern Recognition (2017)

\bibitem{Kaiming2016Deep}
He, K., Zhang, X., Ren, S., Sun, J.: Deep residual learning for image
  recognition. In: The IEEE Conference on Computer Vision and Pattern
  Recognition (2016)

\bibitem{HuSqueeze}
Hu, J., Shen, L., Albanie, S., Sun, G., Wu, E.: Squeeze-and-excitation
  networks. In: The IEEE Conference on Computer Vision and Pattern Recognition
  (2018)

\bibitem{CondenseNet}
Huang, G., Liu, S., van~der Maaten, L., Weinberger, K.Q.: {CondenseNet}: An
  efficient {DenseNet} using learned group convolutions. In: The IEEE
  Conference on Computer Vision and Pattern Recognition (2018)

\bibitem{huang2017densely}
Huang, G., Liu, Z., van~der Maaten, L., Weinberger, K.Q.: Densely connected
  convolutional networks. In: The IEEE Conference on Computer Vision and
  Pattern Recognition (2017)

\bibitem{HuangDeep}
Huang, G., Sun, Y., Liu, Z., Sedra, D., Weinberger, K.: Deep networks with
  stochastic depth. In: European Conference on Computer Vision (2016)

\bibitem{isola2017imagetoimage}
Isola, P., Zhu, J.Y., Zhou, T., Efros, A.A.: Image-to-image translation with
  conditional adversarial networks. In: The IEEE Conference on Computer Vision
  and Pattern Recognition (2017)

\bibitem{Johnson2016Perceptual}
Johnson, J., Alahi, A., Fei-Fei, L.: Perceptual losses for real-time style
  transfer and super-resolution. In: European Conference on Computer Vision
  (2016)

\bibitem{KimAccurate}
Kim, J., Lee, J.K., Lee, K.M.: Accurate image super-resolution using very deep
  convolutional networks. In: The IEEE Conference on Computer Vision and
  Pattern Recognition (2016)

\bibitem{Kim_2016_DRCN}
Kim, J., Lee, J.K., Lee, K.M.: Deeply-recursive convolutional network for image
  super-resolution. In: The IEEE Conference on Computer Vision and Pattern
  Recognition (2016)

\bibitem{kingma2014adam}
Kingma, D.P., Ba, J.: Adam: A method for stochastic optimization. ArXiv
  preprint: 1412.6980  (2014)

\bibitem{LapSRN}
Lai, W.S., Huang, J.B., Ahuja, N., Yang, M.H.: Deep {Laplacian} pyramid
  networks for fast and accurate super-resolution. In: The IEEE Conference on
  Computer Vision and Pattern Recognition (2017)

\bibitem{Christian2017Photo}
Ledig, C., Theis, L., Huszar, F., Caballero, J., Shi, W.: Photo-realistic
  single image super-resolution using a generative adversarial network. In: The
  IEEE Conference on Computer Vision and Pattern Recognition (2017)

\bibitem{li2019expectationmaximization}
Li, X., Zhong, Z., Wu, J., Yang, Y., Lin, Z., Liu, H.: Expectation-maximization
  attention networks for semantic segmentation. In: The IEEE International
  Conference in Computer Vision (2019)

\bibitem{Li2019}
Li, Z.: Image super-resolution using attention based {DenseNet} with residual
  deconvolution. ArXiv preprint: 1907.05282  (2019)

\bibitem{Lin2014Microsoft}
Lin, T.Y., Maire, M., Belongie, S., Hays, J., Zitnick, C.L.: Microsoft {COCO}:
  Common objects in context. In: European Conference on Computer Vision (2014)

\bibitem{LiuAn}
Liu, Y., Wang, Y., Li, N., Cheng, X., Zhang, Y., Huang, Y., Lu, G.: An
  attention-based approach for single image super resolution. In: International
  Conference on Pattern Recognition (2018)

\bibitem{liu2019image}
Liu, Z.S., Wang, L.W., Li, C.T., Siu, W.C., Chan, Y.L.: Image super-resolution
  via attention based back projection networks. In: The IEEE International
  Conference on Computer Vision (2019)

\bibitem{LiuSeg}
Liu, Z., Li, X., Luo, P., Change~Loy, C., Tang, X.: Deep learning {Markov}
  random field for semantic segmentation. IEEE Transactions on Pattern Analysis
  and Machine Intelligence  \textbf{40}(8),  1814--1828 (2018)

\bibitem{OktayAttention}
Oktay, O., Schlemper, J., Folgoc, L.L., Lee, M., Heinrich, M., Misawa, K.,
  Mori, K., Mcdonagh, S., Hammerla, N.Y., Kainz, B.: Attention {U-Net}:
  Learning where to look for the pancreas. In: International Conference on
  Medical Imaging with Deep Learning (2018)

\bibitem{PathakEfficient}
Pathak, H.N., Li, X., Minaee, S., Cowan, B.: Efficient super resolution for
  large-scale images using attentional {GAN}. In: The IEEE International
  Conference on Big Data (2018)

\bibitem{enhancenet}
Sajjadi, M.S.M., Scholkopf, B., Hirsch, M.: {EnhanceNet}: Single image
  super-resolution through automated texture synthesis. In: The IEEE
  International Conference on Computer Vision (2017)

\bibitem{SimonyanVGG}
Simonyan, K., Zisserman, A.: Very deep convolutional networks for large-scale
  image recognition. Computer Science  (2014)

\bibitem{Soh_2019_CVPR_NatSR}
Soh, J.W., Park, G.Y., Jo, J., Cho, N.I.: Natural and realistic single image
  super-resolution with explicit natural manifold discrimination. In: The IEEE
  Conference on Computer Vision and Pattern Recognition (2019)

\bibitem{Flickr}
Timofte, R., Agustsson, E., Gool, L.V., Yang, M.H., Guo, Q.: {NTIRE} 2017
  challenge on single image super-resolution: Methods and results. In: The IEEE
  Conference on Computer Vision and Pattern Recognition Workshops (2017)

\bibitem{WangResidual}
Wang, F., Jiang, M., Qian, C., Yang, S., Li, C., Zhang, H., Wang, X., Tang, X.:
  Residual attention network for image classification. In: The IEEE Conference
  on Computer Vision and Pattern Recognition (2017)

\bibitem{wang2018sftgan}
Wang, X., Yu, K., Dong, C., Loy, C.C.: Recovering realistic texture in image
  super-resolution by deep spatial feature transform. In: The IEEE Conference
  on Computer Vision and Pattern Recognition (2018)

\bibitem{wang2018esrgan}
Wang, X., Yu, K., Wu, S., Gu, J., Liu, Y., Dong, C., Qiao, Y., Loy, C.C.:
  {ESRGAN}: Enhanced super-resolution generative adversarial networks. In:
  European Conference on Computer Vision Workshops (2018)

\bibitem{Wang2004Image}
Wang, Z., Bovik, A.C., Sheikh, H.R., Simoncelli, E.P.: Image quality
  assessment: From error visibility to structural similarity. IEEE Trans Image
  Process  \textbf{13}(4),  600--602 (2004)

\bibitem{semantic2019}
Wu, X., Lucas, A., L{\'{o}}pez{-}Tapia, S., Wang, X., Kim, Y.H., Molina, R.,
  Katsaggelos, A.K.: Semantic prior based generative adversarial network for
  video super-resolution. In: European Signal Processing Conference (2019)

\bibitem{Zhang2018Self}
Zhang, H., Goodfellow, I., Metaxas, D., Odena, A.: Self-attention generative
  adversarial networks. ArXiv preprint: 1805.08318  (2018)

\bibitem{zhang2018rcan}
Zhang, Y., Li, K., Li, K., Wang, L., Zhong, B., Fu, Y.: Image super-resolution
  using very deep residual channel attention networks. In: European Conference
  on Computer Vision (2018)

\bibitem{Zhou2017Scene}
Zhou, B., Hang, Z., Puig, X., Fidler, S., Barriuso, A., Torralba, A.: Scene
  parsing through {ADE20K} dataset. In: The IEEE Conference on Computer Vision
  and Pattern Recognition (2017)

\bibitem{Zhu2017Be}
Zhu, S., Fidler, S., Urtasun, R., Lin, D., Loy, C.C.: Be your own prada:
  Fashion synthesis with structural coherence. In: The IEEE International
  Conference in Computer Vision (2017)

\bibitem{nas}
Zoph, B., Le, Q.V.: Neural architecture search with reinforcement learning.
  ArXiv preprint: 1611.01578  (2016)

\end{thebibliography}
\end{document}